\documentclass[review]{elsarticle}

\usepackage{lineno,hyperref}
\usepackage{amsmath,graphicx}

\usepackage{amssymb} 
\usepackage{color}
\usepackage{booktabs}
\usepackage{multirow}
\usepackage{mathdots}
\usepackage{algorithm}
\modulolinenumbers[5]

\journal{Journal of \LaTeX\ Templates}

\DeclareMathOperator*{\argmin}{arg\,min}

\begin{document}

\begin{frontmatter}

\title{A novel learning-based frame pooling method for Event Detection}


\author[mymainaddress]{Lan Wang}
\author[mymainaddress]{Chenqiang Gao\corref{mycorrespondingauthor}}\cortext[mycorrespondingauthor]{Corresponding author}
\ead{gaocq@cqupt.edu.cn}
\author[mymainaddress]{Jiang Liu}
\author[mysecondaryaddress]{Deyu Meng}



\address[mymainaddress]{Chongqing Key Laboratory of Signal and Information Processing, Chongqing University of Posts and Telecommunications, Chongqing, China}
\address[mysecondaryaddress]{Institute for Information and System Sciences Faculty of Mathematics and Statistics, Xi'an Jiaotong University, Xi'an, China}

\begin{abstract}
Detecting complex events in a large video collection crawled from video websites is a challenging task. When applying directly good image-based feature representation, e.g., HOG, SIFT, to videos, we have to face the problem of how to pool multiple frame feature representations into one feature representation. In this paper, we propose a novel learning-based frame pooling method. We formulate the pooling weight learning as an optimization problem and thus our method can automatically learn the best pooling weight configuration for each specific event category. Experimental results conducted on TRECVID MED 2011 reveal that our method outperforms the commonly used average pooling and max pooling strategies on both high-level and low-level 2D image features.
\end{abstract}


\end{frontmatter}

\linenumbers

\section{Introduction}\label{sec:intro}

\begin{figure}[!h]
	\centering{\includegraphics[scale=0.5]{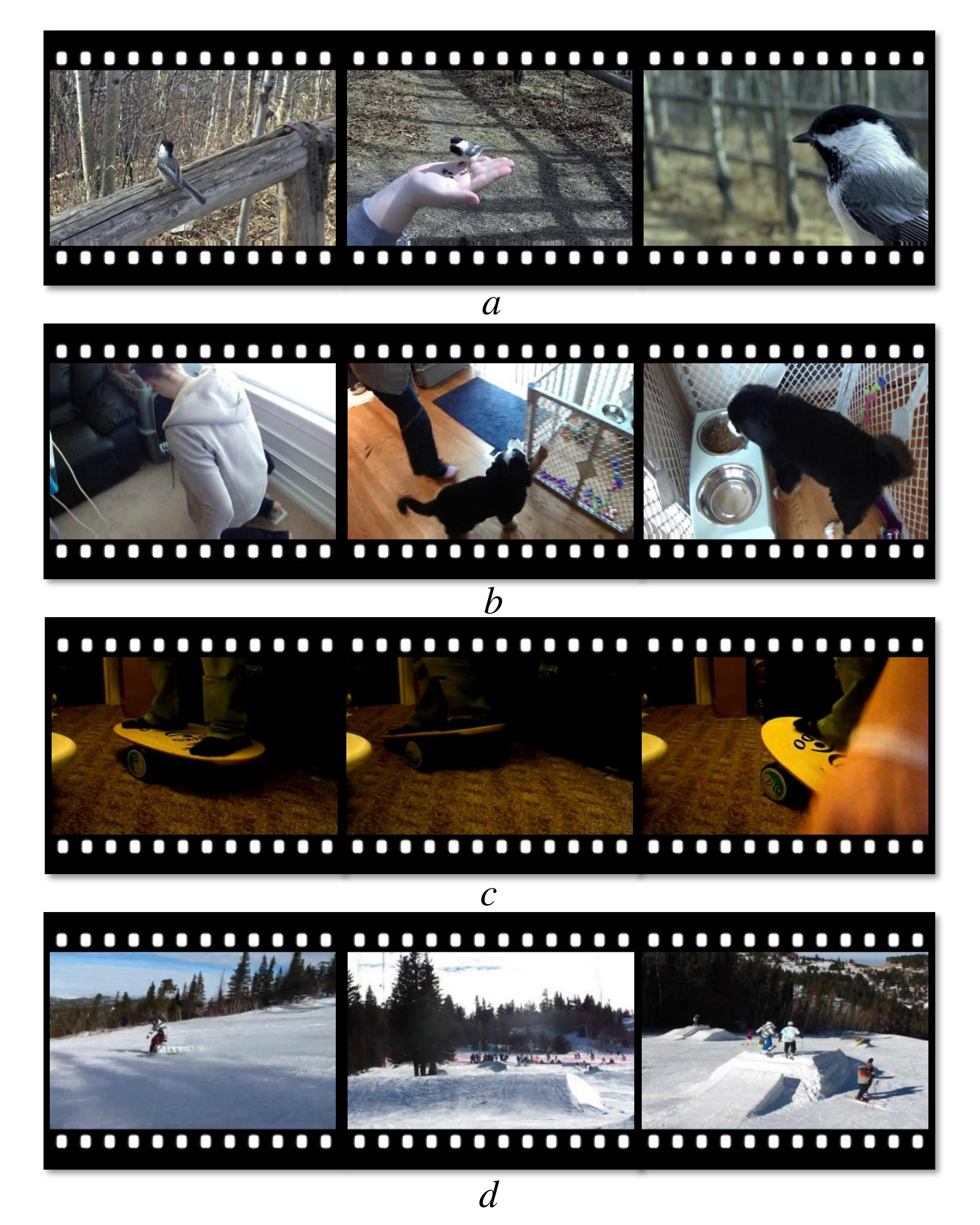}}
	\caption{Example frames of video clips in different events:
		(a)(b) Instances of the  ``Feeding an animal" event. (c)(d) Instances of the ``Attempting a aboard trick" event.}
	\label{fig:event}
\end{figure}

Complex event detection aims to detect events, such as ``marriage proposal'', ``renovating a home'', in a large video collection crawled from video websites, like Youtube. This technique can be extensively applied to Internet video retrieval, content-based video analysis and machine intelligence fields and thus has recently attracted much research attention\cite{tang2012learning,lan2013cmu,gao2014interactive,xu2014discriminative,yang2015novel}. Nevertheless, the complex event detection encounters lots of challenges, mostly because events are usually more complicated and undefinable, possessing great intra-class variations and variable video durations, as compared with traditional concept analysis in constrained video clips, e.g., action recognition\cite{yang2016multi,liu2015coupled}. For example, identical events, as shown in Fig. \ref{fig:event}, are entirely distinct in different videos, with various scenes, animals, illumination and views. Even in the same video, these factors are also changing. The above reasons make event detection far from being applicable to practical use with robust performance.

A large number of methods have been proposed to handle this challenging task\cite{chang2016dynamic,yan2014glocal,yan2015event,ma2014knowledge}. Generally speaking, the video representation is one of the most important components. For many techniques to extract the video representation, namely feature descriptors, have to be carefully designed or selected for good detection performance. Different from images, video clips can be treated as spatial-temporal 3D cuboids. Lots of spatial-temporal oriented feature descriptors have been proposed and been proved effective, such as HOG3D\cite{klaser2008spatio}, MoSIFT\cite{chen2009mosift}, 3DSIFT\cite{scovanner20073} and the state-of-the-art improved Dense Trajectory(IDT)\cite{wang2013dense}. Although these spatial-temporal descriptors can intrinsically describe videos, the 2D image descriptors are still very important for describing videos in the complex event detection community due to two aspects. On one hand, compared with 2D image descriptors, the spatial-temporal feature descriptors usually require larger data storage and higher computational complexity to be extracted and processed. This problem becomes more serious for large scale datasets. On the other hand, the TRECVID Multimedia Event Detection (MED) evaluation track\cite{1178722} of each year, held by NIST, reveals that combining kinds of feature descriptors, including 2D and 3D features, usually outperforms those of using a single feature descriptor\cite{lan2013informedia}.

Profiting from the research development in image representations, a number of good features, including low-level ones of such HOG\cite{dalal2005histograms}, SIFT\cite{lowe1999object}, and high-level features of such Object-bank\cite{li2010object} along with the recently most successful Convolutional Neural Network(CNN) feature\cite{donahue2013decaf} can be directly applied to describe the video. The commonly used strategy is to extract the feature representation for each frame or selected key frames of the video (we will use \emph{frame} hereinafter) and then pool all feature representations into one representation with $average\ pooling$ or $max\ pooling$\cite{boureau2010theoretical}. While the max pooling just uses the maximum response of all frames for each feature component, the average pooling uses their average value. It is hard to say which one of these two pooling strategies is better. Sometimes, average pooling is better than max pooling and vice versa. The performance heavily depends on the practical application or datasets. The actual strategy is manually choosing the better one through experiments conducted on a validation set. Therefore, intuitively, here comes two questions: 1) can we automatically choose the better one between the two previous pooling strategies? 2) is there any pooling method superior to these two strategies?

To answer these two questions mentioned above, we propose a novel learning-based frame pooling method. We notice that when human beings observe different events, they usually have different attention on various frames, i.e., the pooling weight for a particular event is inconsistent with the others. This pheneomenon inspires us to adaptively learn the optimal pooling way from data. In other words, our approach can automatically derive the best pooling weight configuration for each specific event category. To this end, we design an alternative search strategy, which embeds the optimization process for frame pooling weight and classifier parameters into an unifying optimization problem.
Experimental results conducted on TRECVID MED 2011 reveal that our learning-based frame pooling method outperforms the commonly used average pooling and max pooling strategies on both high-level and low-level 2D image features.

The rest part of this paper is organized as following. In Section \ref{sec:method}, we present our proposed methodology for video description task. Section \ref{sec:exp} shows the experimental results with various low-level and high-level features. The conclusion is finally given in Section \ref{sec:conclusion}.

\section{The proposed method}\label{sec:method}

\subsection{ Overview of our framework}\label{sec:overview}
\begin{figure}[!h]
	\centering{\includegraphics[scale=0.4]{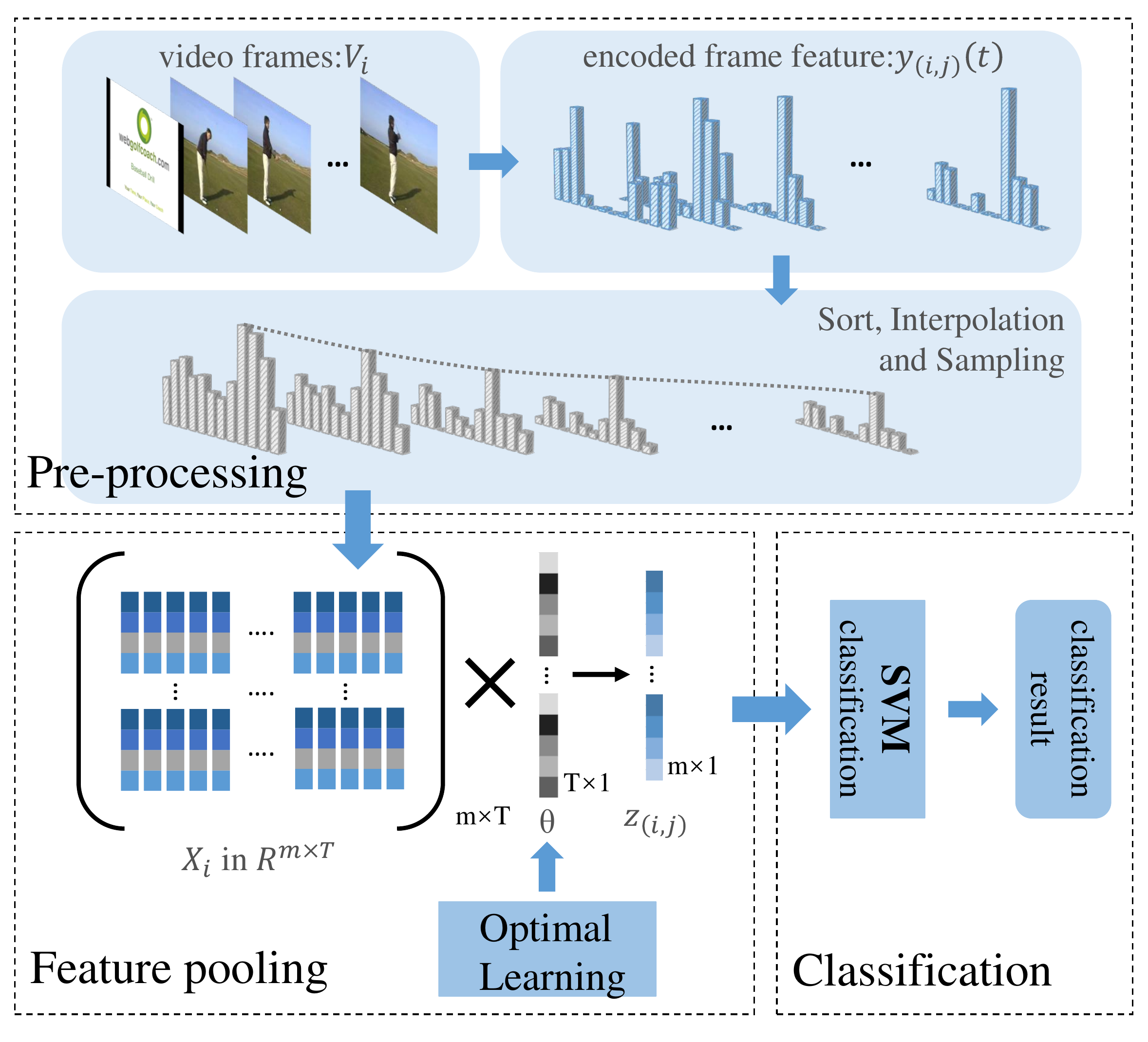}}
	\caption{
		The framework of our method. }
	\label{fig:pipeline}
\end{figure}

In this section, we briefly describe the learning-based frame pooling method. The proposed algorithm consists of  three main modules: pre-processing, feature pooling  and classification, as shown in Fig. \ref{fig:pipeline}.

During the pre-processing stage, we extract the features of all frames and then sort all components in descent order. Then, the lagrange interpolation and sampling operations are conducted on each video with different frames, to get fixed number features. In pooling stage, we pool
the features with learned optimal pooling weights learned by our optimal pooling method described below. Finally, a classifier is employed to obtain the classification results.

\subsection{Pre-processing}\label{sec:interp}

Our goal is to learn an uniform pooling weight setting for each specific event. However, the number of frames extracted from videos containing events are different due to different video durations or frame sampling methods. To address this problem, the interpolation operation is adopted.

Given a video clip $V_{i}$ with $T_{i}$ frames, we can get $T_{i}$ encoded feature vectors $y_{(i,j)}(t), t\in(1,2,3,\cdots,T_{i}), j \in(1,2,3,\cdots,m) $. Here $m$ is the dimension of the feature in each frame. First, we construct a Lagrange interpolation function $\tilde{f_{i,j}}(u)$ for the $j^{th}$ feature component as following:
\begin{equation}
\tilde{f_{i,j}}\left( u\right)=\sum^{T_{i}}_{t=1}\dfrac{\prod ^{t-1}_{k=1}\left(u-k\right)\prod^{T_{i}}_{k=t+1}\left( u-k\right) }{\prod^{t-1}_{k=1}\left(t-k\right) \prod^{T_{i}}_{k=t+1}\left( t-k \right) }y_{i,j}\left( t\right),
\label{equ:4}
\end{equation}
where $\tilde{f_{i,j}}(u)$ can fit all the responses at each time (frame) $u$ in the original video clip. With the interpolated functions for all feature components, we can re-sample a fixed number of the feature representations. Thus, the videos with various durations are eventually able to re-normalized into ones with the same number $T$ of feature representations. However, we would encounter the ``over-fitting" problem if directly conducting interpolating operation on the original encoded features. This is due to the fact that the original feature components may varies greatly even between consecutive frames and hence will cause the corresponding interpolation function to vary dramatically in the feature space. This would produce potential noise data. For the sake of alleviating this problem, we sort independently all features for each component in descent order before constructing the Lagrange interpolation function. In this way, the interpolation function will tend to gradually decreasing in the feature space, we denote it as $f_{i,j}\left( u\right)$. Later, we sample along the temporal axis for the $j^{th}$ feature component with  $f_{i,j}\left( u\right)$, denoted as $\overline{x}_{i,j}$:
\begin{equation}
\overline {x}_{i,j}=\left\{ f_{i.j}\left( t^{i}_{k}\right) \right\}, k\in(1,2,3,\cdots,T),
\label{equ:5}
\end{equation}
where $t^{i}_{k}=1+\left( k-1\right) \dfrac {T_{i}-1}{T-1}$, are the re-sampling points on the interpolated function. For a given video clip, we combine all sampled feature vectors together into a new feature matrix, denoted as $ X_{i}=(\overline{x}_{i,1},\overline{x}_{i,2},\overline{x}_{i,3}...\overline{x}_{i,m})^{T} \in \mathbb{R} ^{m\times T} $.

\subsection{Formulation}\label{sec:formulation}

Given $n$ training samples $(X_{i},y_i)(i=1,2,\cdots,n)$, where the $X_{i}$ is the feature matrix obtained by Section \ref{sec:interp} and $y_i$ is the sample label, our goal is to learn a weight parameter to pool the feature matrix $X_{i}$ into a single feature vector. Actually, for both average and max pooling methods, the pooling operation is done independently for each feature component. Intuitively, we should learn an independent weight vector $\theta^j(j=1,\cdots,m)$ for each component. However, this would make the model too complex to be learned effectively. Instead, we learn a single weight vector $\theta$ for all components. Namely, we pool the features using the same weight vector for all feature components as $X_{i}\theta$. Because our interpolation function $f_{i,j}$ will perform a decreasing property in feature space, we can easily know that the cases of $\theta=(1/T,\cdots,1/T)$ and $\theta=(1,0,0,\cdots,0)$ approximately correspond to average and max pooling strategies, respectively. Furthermore, the medium and minimum pooling strategies can also be approximately viewed as two specific cases, where $\theta=(0,\cdots,1,\cdots,0)$(1 is located in the middle position of the vector) and $\theta=(0,0,\cdots,1)$, respectively. Nevertheless, our goal is to learn an optimal pooling strategy for each event. To this end, the problem of pooling parameter $\theta$ learning is formulated as the following optimization problem:



\begin{equation}
\left\{
\begin{aligned}
\min _{w,b,\theta }\sum ^{n}_{i=1}\left( 1-y_{i}\left( w^{T}X_{i}\theta +b\right) \right) _{+}+\dfrac {1}{2}w^{T}w,\\
s.t\ \ \theta \geq 0, \sum ^{T}_{k=1}\theta _{k}=1,
\end{aligned}
\right.
\label{equ:6}
\end{equation}
where $(\cdot)_{+}=max(0,\cdot)$ means the hinge-loss in the loss function. Our model intends to minimize the objective function over $w,b$, which are the parameters of the hyperplane in the SVM classifier, along with our additional pooling parameter $\theta$.


\subsection{Solution}\label{sec:solution}

In order to solve the parameters of $w,b,\theta$ in the model (\ref{equ:6}) above, an alternative search strategy is employed. In general, our alternative search strategy can be viewed as an iteration approach with two steps in each round. The first step in each iteration is to update $w,b$ with fixed $\theta$ by solving the following sub-optimization problem:

\begin{equation}
(w,b)=\argmin_{w,b }\sum ^{n}_{i=1}\left( 1-y_{i}\left( w^{T}X_{i}\theta +b\right) \right) _{+}+\dfrac {1}{2}w^{T}w.
\label{equ:7}
\end{equation}
Here, we initialize $\theta$ with random values with constraint that $\theta \geq 0, \sum ^{T}_{k=1}\theta _{k}=1$.
Equation (\ref{equ:7}) is the standard formulation of a linear SVM problem and therefore can be solved via off-the-shelf tools like libsvm\cite{chang2011libsvm}.

The second step in an iteration is to search $\theta$ by fixing the $w,b$ obtained by the first step. This step actually iteratively updates an optimal pooling manner under current model parameter $w,b$:

\begin{equation}
\theta=\argmin_{\theta }\sum ^{n}_{i=1}\left( 1-y_{i}\left( w^{T}X_{i}\theta +b\right) \right)_{+} \ \  s.t\ \ \theta \geq 0, \sum ^{T}_{k=1}\theta _{k}=1.
\label{equ:8}
\end{equation}
Directly solving this optimization problem would be very complex because the hinge loss and the constraints on $\theta$ make it a non-convex function. In this degree, a transformation of the above optimization problem needs to be conducted by relaxing the convex property. For each particular sample in the training set, we first introduce an upper bound $\varepsilon_{i}$, measuring the corresponding upper bound of the classification error in the SVM model. According to the hinge loss property, the following two conditions are obtained:
\begin{equation}
\varepsilon_{i} \geq 1-y_{i}\left( w^{T}X_{i}\theta +b\right),
\label{equ:8}
\end{equation}

\begin{equation}
\varepsilon_{i} \geq 0.
\label{equ:9}
\end{equation}
Eliminating the hinge loss using these conditions gives the reformulation of the optimization problem:

\begin{equation}
\theta=\argmin_{\theta,\varepsilon }\sum ^{n}_{i=1}\varepsilon_{i} \ \ s.t\ \ \theta \geq 0, \sum ^{T}_{k=1}\theta _{k}=1, y_{i}\left( w^{T}X_{i}\theta +b\right) \geq 1-\varepsilon_{i}, \varepsilon_{i} \geq 0,
\label{equ:10}
\end{equation}
We can further transform equation(\ref{equ:10}) into a constrained linear programming(LP) problem by defining $\alpha=\left[ \theta,\varepsilon_{1},\ldots,\varepsilon_{d}\right]$, $\rho=\left[ 0,\ldots,0,1,\ldots,1\right]$, $ \sigma=\left[ 1,\ldots,1,0,\ldots,0\right]$, $ \eta_{i}=\left[y_{i} w^{T}X_{i}, e_{i}\right]$. Then, equation(\ref{equ:10})  can be rewritten as follows:

\begin{equation}
\alpha=\argmin_{\alpha} \rho\alpha^{T} \ \ s.t\ \  \alpha \geq 0,\sigma\alpha^{T}=1,\eta_{i}\alpha^{T}  \geq 1-y_{i}b,
\label{equ:11}
\end{equation}
where $d$ denotes the number of video clips. The number of zeros in $\rho$ equals to the length of vector $\theta$, and then follows $d$ of ones. In other words, $\rho$ plays a role as an selection variable which picks out $\varepsilon$ terms in $\alpha$. On the other hand, $\delta$ oppositely selects out the $\theta$, which contains $\|\theta\|$ numbers of ones and $d$ of zeros. Equation(\ref{equ:11}) is a classical linear programming model, which can be optimize using existing tools.

In this way, the overall objective function can be minimized with expected convergence by iteratively searching for $w,b$ and $\theta$, respectively. The overall algorithm is illustrated in Algorithm \ref{alg1}.

\begin{algorithm}[htbp]
	\caption{Alternative search strategy to obtain optimum $w,b,\theta$}
	\textbf{Input:} $X_{i}$, $y_{i}$(the training set feature matrices and labels),  \\
	\textbf{Output:} learned parameter $w,b,\theta$
	\begin{enumerate}
		\item Initialize $\theta^{(1)}$ with random values, $s.t.\ \ \theta^{(1)} \geq 0, \sum ^{T}_{j=1}\theta^{(1)}_{j}=1$;
		\item \textbf{\textit{for k:=2 to N}}
		\begin{enumerate}
			\item Fixing $\theta^{(k-1)}$ and updating $w^{(k)}$,$b^{(k)}$: \\
			$(w^{(k)},b^{(k)})=\argmin_{w,b }\sum ^{n}_{i=1}\left( 1-y_{i}\left( w^{T}X_{i}\theta^{(k-1)} +b\right) \right)_{+}+\dfrac {1}{2}w^{T}w$;
			\item Update $\alpha^{(k)}$:\\
			$\alpha^{(k)}=\argmin_{\alpha} \rho\alpha^{T} \ \ s.t\ \ \alpha \geq 0,\sigma\alpha^{T}=1,\eta_{i}\alpha^{T}  \geq 1-y_{i}b^{(k)}$;
			\item Obtain $\theta^{(k)}$ according to $\alpha^{(k)}$;
		\end{enumerate}
		\textbf{\textit{end for}}
		\item Return $w^{(N)}$, $b^{(N)}$ and $\theta^{(N)}$.
	\end{enumerate}
	\label{alg1}
\end{algorithm}

\section{Experiments}\label{sec:exp}

We evaluate our proposed model on the public large scale TRECVID MED2011 dataset\cite{1178722} with both low-level features: HOG\cite{dalal2005histograms}, SIFT\cite{lowe1999object}, and high-level features: Object Bank-based feature\cite{li2010object} and CNN-based feature\cite{donahue2013decaf}. We adopt the most popular pooling methods of the \emph{max} and \emph{average} poolings as the baseline methods for comparison.

\subsection{Dataset and evaluation metric}
The TRECVID MED 2011 development set\cite{1178722} is used to evaluate our method. It contains more than 13,000 video clips over 18 different kinds of events and background classes, which provides us with real life web video instances consisting of complex events under different scenes lasting from a few seconds to several minutes. The specific events are listed in Table \ref{tab:MED}. We follow the original evaluation metric along with the pre-defined training/test splits of MED 2011 development set. In the pre-processing stage, we empirically interpolate each video clips into $T=20$ frames. Besides, each learning-based frame pooling model for individual event class is trained with 100 times of iteration, which enables the objective function to be minimized to convergent. Finally, the average precision(AP) and the mean average precision(mAP) values are used as the evaluation metrics for different pooling approaches. Mean average precision is defined as the mean AP over all events.

\begin{table}[htbp]\scriptsize
	\centering
	\caption{18 events of TRECVID MED 11.}
	\begin{tabular}{cccc}
		\toprule
		& \textbf{Event} &       & \textbf{Event} \\
		\midrule
		E001  & Attempting a aboard trick & E010  & Grooming an animal \\
		E002  & Feeding an animal & E011  & Making a sandwich \\
		E003  & Landing a fish & E012  & Parade \\
		E004  & Working on a woodworking project & E013  & Parkour \\
		E005  & Wedding ceremony & E014  & Repairing an appliance \\
		E006  & Birthday party & E015  & Working on a sewing project \\
		E007  & Changing  a vehicle tire & P001  & Assembling shelter \\
		E008  & Flash mob gathering & P002  & Batting a run \\
		E009  & Getting a vehicle unstuck & P003  & Making a cake \\
		\bottomrule
	\end{tabular}%
	\label{tab:MED}%
\end{table}%

\subsection{Results on low-level features}
We use the off-the-shelf toolkit \emph{VLFeat}\cite{vedaldi2010vlfeat} to extract HOG and SIFT features with standard configurations for each frame. Here the SIFT descriptors are densely extracted. Then the Bag-of-Words method is employed to encode the raw features from each frame into a 100 dimensional vector. The results are listed in Table \ref{tab:lowlevel}.

\begin{table}[htbp]\scriptsize
	\centering
	\caption{The AP comparison among average pooling, max pooling and our optimal pooling method for low-level features on TRECVID MED11 dataset.}
	\label{tab:lowlevel}%
	\begin{tabular}{cccc|ccc}
		\toprule
		\multirow{2}[4]{*}[9pt]{\textbf{Event ID}} & \multicolumn{3}{c}{\textbf{HOG}} & \multicolumn{3}{c}{\textbf{SIFT}} \\
		\midrule
		\textbf{Method} &\textbf{Average}  & \textbf{Max} & \textbf{Ours} & \textbf{Average} & \textbf{Max} & \textbf{Ours} \\
		E001 & 0.407  & 0.435  & \textbf{0.457 } & 0.270 & 0.275  & \textbf{0.298 } \\
		E002 & 0.302  & 0.320  & \textbf{0.369 } & 0.207  & 0.217  & \textbf{0.223 } \\
		E003 & 0.527  & 0.511  & \textbf{0.586 } & 0.290 & 0.252  & \textbf{0.294 } \\
		E004 & 0.279  & \textbf{0.307 } & 0.285  & 0.140 & \textbf{0.158 } & 0.130  \\
		E005 & 0.184  & \textbf{0.217 } & 0.189  & 0.142 & \textbf{0.185 } & 0.165  \\
		E006 & 0.179  & 0.175  & \textbf{0.220 } & 0.098 & \textbf{0.145 } & 0.138  \\
		E007 & 0.083  & \textbf{0.112 } & 0.102  & 0.081 & 0.076  & \textbf{0.082 } \\
		E008 & 0.162  & 0.269  & \textbf{0.325 } & 0.197 & 0.181  & \textbf{0.201 } \\
		E009 & 0.327  & 0.357  & \textbf{0.362 } & 0.103 & 0.149  & \textbf{0.180 } \\
		E010 & 0.151  & 0.136  & \textbf{0.180 } & 0.113 & \textbf{0.151 } & 0.125  \\
		E011 & 0.082  & 0.080  & \textbf{0.096 } & 0.085 &  0.071  & \textbf{0.112 } \\
		E012 & 0.107  & 0.144  & \textbf{0.153 } & 0.141 &  0.206  & \textbf{0.216 } \\
		E013 & 0.110  & 0.126  & \textbf{0.130 } & \textbf{0.107} &  0.091  & 0.104  \\
		E014 & 0.192  & 0.177  & \textbf{0.233 } & 0.150 &  \textbf{0.177 } & 0.154  \\
		E015 & 0.097  & 0.104  & \textbf{0.157 } & 0.185 &  0.180  & \textbf{0.195 } \\
		P001 & 0.123  & \textbf{0.162 } & 0.147  & 0.105 &  0.129  & \textbf{0.130 } \\
		P002 & 0.350  & 0.379  & \textbf{0.424 } & 0.362 &  0.344  & \textbf{0.362 } \\
		P003 & 0.057  & 0.066  & \textbf{0.117 } & 0.058 &  0.044  & \textbf{0.065 } \\
		\textbf{mAP} & 0.207  & 0.226  & \textbf{0.252 } & 0.158 & 0.168  & \textbf{0.176 } \\
		\bottomrule
	\end{tabular}%
\end{table}%

From Table \ref{tab:lowlevel}, it can be obviously observed that our method is effective on most events for both HOG and SIFT features. For the HOG descriptor, our model leads to apparent AP improvements on 14 out of 18 events, and  our learning-based method outperforms the max and average pooling strategies by 0.026 and 0.045 in mAP, respectively. As to the SIFT descriptor, the APs of overall 12 out of 18 events are improved by our method and our method outperforms the max and average pooling strategies by 0.008 and 0.018 in mAP, respectively. It is worth noting that it is very hard to improve mAP, even by 0.01 since the TRECVID MED11 is a very challenging dataset.
\begin{table}[hbtp]\scriptsize
	\centering
	\caption{Comparisons among different methods for high-level features on TRECVID MED11 dataset.}
	\begin{tabular}{cccc|ccc}
		\toprule
		\multirow{2}[4]{*}[9pt]{\textbf{Event ID}}  & \multicolumn{3}{c}{\textbf{Max-OB}} & \multicolumn{3}{c}{\textbf{CNN 128d}} \\
		\midrule
		\textbf{Method} & \textbf{Average} & \textbf{Max} & \textbf{Ours} & \textbf{Average} & \textbf{Max} & \textbf{Ours} \\
		E001 & 0.443 & \textbf{0.445 } & 0.436 & 0.645 & 0.653  & \textbf{0.654 } \\
		E002 & 0.321 & 0.338  & \textbf{0.403 } & \textbf{0.394} & 0.388  & \textbf{0.394 } \\
		E003 & 0.191 & 0.184  & \textbf{0.216 } & 0.746 & 0.745  & \textbf{0.747} \\
		E004 & 0.128 & 0.129  & \textbf{0.168 } & \textbf{0.820} & 0.818 & 0.813  \\
		E005 & \textbf{0.153} & 0.151 & 0.131  & 0.502 & \textbf{0.590 } & 0.581  \\
		E006 & 0.370 & 0.368  & \textbf{0.384 } & 0.387 & 0.389  & \textbf{0.389} \\
		E007 & 0.077 & 0.075  & \textbf{0.132 } & 0.333 & 0.323  & \textbf{0.337} \\
		E008 & 0.120 & 0.121  & \textbf{0.244 } & 0.423 & 0.446  & \textbf{0.461 } \\
		E009 & 0.318 & 0.320  & \textbf{0.362 } & 0.632 & 0.627  & \textbf{0.636 } \\
		E010 & 0.124 & \textbf{0.127 } & 0.119  & 0.214 & 0.269  & \textbf{0.303 } \\
		E011 & 0.186 & 0.243  & \textbf{0.268 } & 0.250 &  0.249  & \textbf{0.252 } \\
		E012 & 0.178 & \textbf{0.211 } & 0.183  & 0.371 & 0.425  & \textbf{0.425 } \\
		E013 & 0.123 & 0.110  & \textbf{0.125 } & 0.309 & \textbf{0.327}  & 0.326  \\
		E014 & 0.175 & \textbf{0.246 } & 0.169  & 0.384  & 0.381  & \textbf{0.384 } \\
		E015 & 0.210 & 0.191  & \textbf{0.219 } & 0.410  & 0.410  & \textbf{0.422 } \\
		P001 & 0.201 & 0.172  & \textbf{0.203 } & 0.426  & 0.453  & \textbf{0.447 } \\
		P002 & 0.211 & 0.198  & \textbf{0.224 } & 0.851  & \textbf{0.956 } & 0.949  \\
		P003 & 0.118 & 0.133  & \textbf{0.144 } & 0.224 & 0.219  & \textbf{0.227 } \\
		\textbf{mAP} & 0.203 & 0.209 & \textbf{0.229}  & 0.484  & 0.481  & \textbf{0.486 } \\
		\bottomrule
	\end{tabular}%
	\label{tab:highlevel}%
\end{table}%

\subsection{Results on high-level features}
We test two kinds of high-level features: CNN-based feature and Object Bank-based feature. When it comes to the CNN-based feature, we directly employ the vgg-m-128 network\cite{Chatfield14}, pre-trained on ILSVRC2012 dataset, to extract feature on each single frame. In detail, we use the 128 dimensional fully connected layer feature as the final feature descriptor, denoted as ``CNN 128d". The Object Bank-based descriptor is a combination of several independent ``object concept" filter responses, where We pre-train 1,000 Object filters on the ImageNet dataset \cite{deng2012imagenet}. For each video frame, we employ the maximum response value for each filter as the image-level filter response. Thus, each frame is represented with a 1,000 dimensional descriptor, denoted as ``Max-OB". The experiment results are listed shown in Table \ref{tab:highlevel}.

Basically, consistent with the low-level feature descriptors, our learning-based pooling method is also effective for both two high-level features on most events. For some specific events, the improvements are large using our method. For example, in \emph{E008}, the event of ``Flash mod gathering" for object bank-based feature, our method improves the AP by more than 0.12 compared with average and max pooling methods. Averagely, our method has an improvement of around 0.02 in mAP compared to baseline methods for object bank-based feature, while around 0.002 in mAP for CNN-based feature.

From Table \ref{tab:lowlevel} and \ref{tab:highlevel}, we can see that it is hard to determine which one of the baseline methods is better. Their performances rely heavily on the feature descriptors and event types. In contrast, our method performs the best in most cases(and in average).

\subsection{Visualization of learned pooling parameters}
\begin{figure}[!h]
	\centering{\includegraphics[scale=.35]{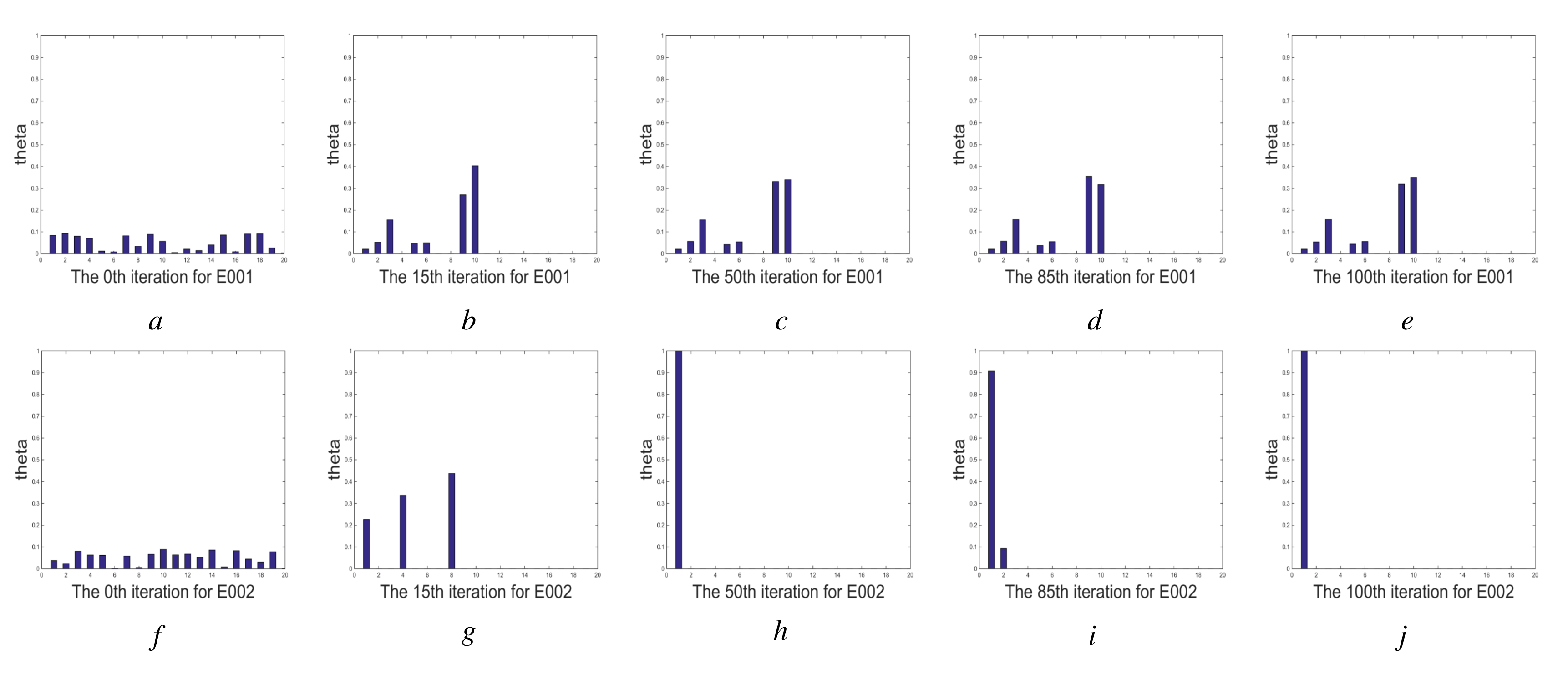}}
	\caption{The learned weights of \emph{E001} and \emph{E002} in iterations, for HOG feature: (a) - (e) The learned weights in the 0th, 15th, 50th, 85th and 100th iteration for E001. (f) - (j) The learned weights in the 0th, 15th, 50th, 85th and 100th iteration for E002.}
	\label{fig:fig2}
\end{figure}

To get an intuitive observation of the optimal weights, we plot the value of $\theta$ in all iterations which are illustrated in Algorithm \ref{alg1}, step 1 and 2, and observe the variation trend, as shown in Fig \ref{fig:fig2}. In this experiment, we select  \emph{E001} and \emph{E002}, the event of ``Attemping a aboard trick" and ``Feeding an animal", for HOG feature and sample several iterations representing the whole process. From the results, we can see that the weights in some components increase quickly at first, and then reach stable high values. In  \emph{E002}, the weights gradually concentrate in the first component, which shows the validity of max pooling. In  \emph{E001}, as can be seen, features in the 9th and the 10th ranking order play a leading role, which illustrate that the most important weight is not always appearing in the order of the max value(max pooling), the min value(min pooling), or the average of all orders(average pooling).

\begin{figure}[!h]
	\centering{\includegraphics[scale=.4]{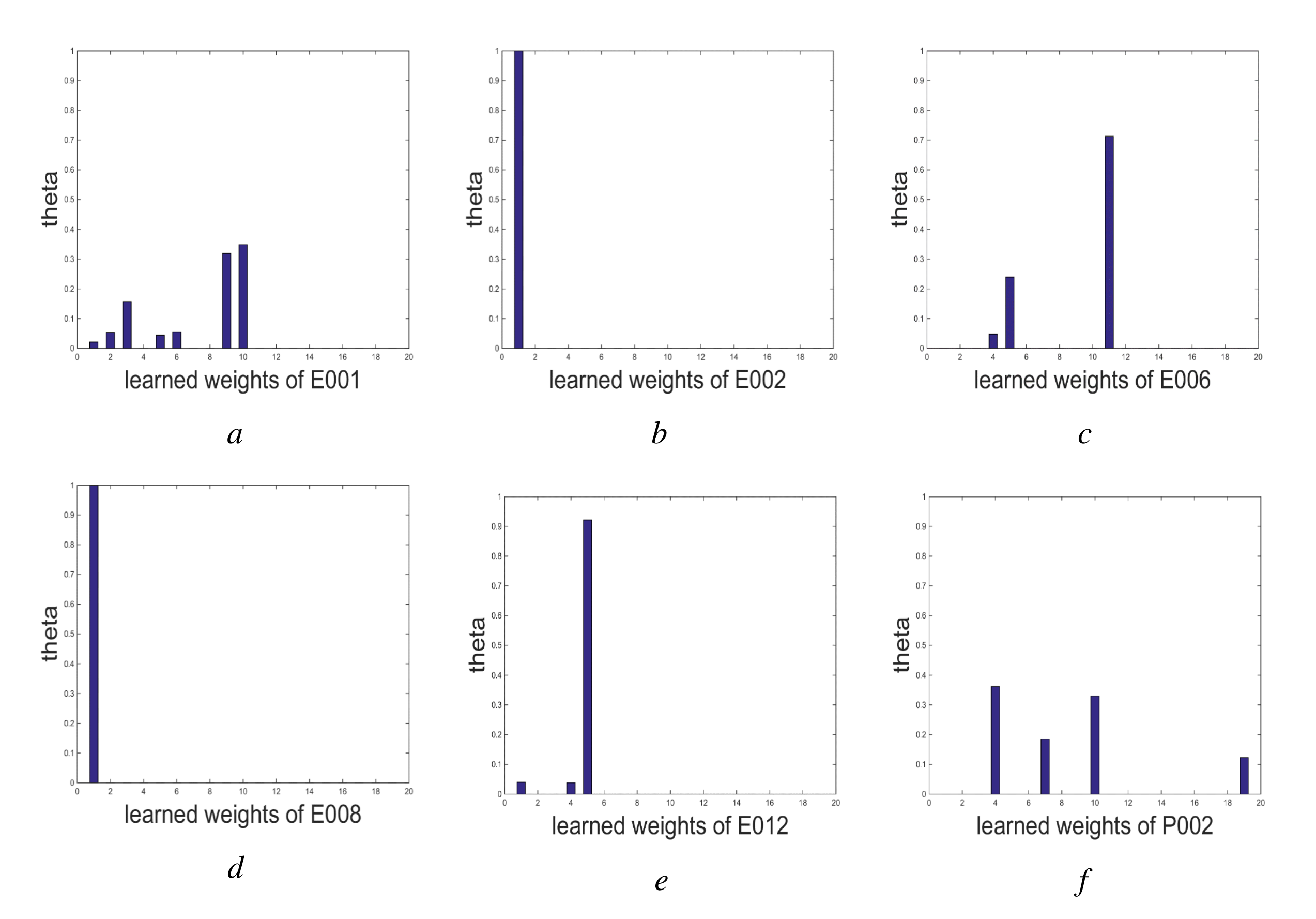}}
	\caption{The learned weights in different events.}
	\label{fig:fig3}
\end{figure}

To observe the influence of event classes, we next compare the learned weights on different events. We select several events
, calculate and visualize their weights of the final iterations for HOG feature, as shown in Fig \ref{fig:fig3}. 
The events \emph{E002} and \emph{E008} are ``Feeding an animal" and ``Flash mob gathering", of which the features with max values carries the greatest significance. They get similar weights which have a tendency of max pooling (notice that the weights of their first components are not equal to one, other components also get values which are too small to display visually). Coincidentally, for HOG feature, max pooling yields better mAP than average pooling. In these cases, the optimal weights outperform max pooling just because of the tiny values in components except the max one(i.e. the first component). The distributions of events \emph{E001}, \emph{E008}, \emph{E012} and \emph{P002} are scattered, 
reflecting that arbitrary components are indispensable. From the result, we can see the weights of events are various, which illustrates that the focus of video clip changes with event classes.

\section{Conclusion}
\label{sec:conclusion}
In this paper, we propose a learning-based frame pooling method to address the complex event detection task. Compared with commonly used average pooling and max pooling approaches, our method can automatically derive the pooling weight among frames for each event category. Through visualization, the learned weights reveal that weight distributions differ in all event categores. Even more, in each event, trivial weight components are also non-ignorable. Experimental results demonstrate that our approach is more effective and robust for both low-level and high-level image descriptors compared with traditional pooling methods.

\section*{References}

\bibliography{mybibfile}

\end{document}